\documentclass[runningheads]{llncs}
\usepackage[T1]{fontenc}
\usepackage{graphicx,verbatim}
\usepackage{amsmath}
\usepackage{amssymb}
\usepackage{booktabs}

\newcommand{\fig}[1]{Fig.~\ref{fig:#1}}
\newcommand{\figs}[2]{Figures~(\ref{fig:#1}-\ref{fig:#2})}
\newcommand{\tab}[1]{Tab.~\ref{tab:#1}}

\usepackage{hyperref}
\usepackage{color}

\usepackage{xspace}

\newcommand{\update}[1]{#1}  
\newcommand{\camera}[1]{#1}

\newcommand{\latinphrase}[1]{\textit{#1}}  
\newcommand{\etal}{\latinphrase{et~al.}\xspace}
\newcommand{\ie}{\latinphrase{i.e.}\xspace}

\begin{document}
\title{Skip priors and add graph-based anatomical information, for point-based Couinaud segmentation}
%

\author{Xiaotong Zhang, Alexander Broersen, Gonnie CM van Erp,\\ Silvia L. Pintea, Jouke Dijkstra*}  

\institute{Radiology department, Leiden University Medical Center,\\ Albinusdreef 2, Leiden, 2333 ZA, The Netherlands. \\
    *Corresponding author: \email{j.dijkstra@lumc.nl}}
\maketitle              

\begin{abstract}
The preoperative planning of liver surgery relies on Couinaud segmentation from computed tomography (CT) images, to reduce the risk of bleeding and guide the resection procedure.
\update{Using $3$D point-based representations, rather than voxelizing the CT volume, has the benefit of preserving the physical resolution of the CT.}
However, point-based representations need prior knowledge of the liver vessel structure, which is time consuming to acquire.
Here, we propose a point-based method for Couinaud segmentation, without explicitly providing the prior liver vessel structure.
To allow the model to learn this anatomical liver vessel structure, we add a graph reasoning module on top of the point features. 
This adds implicit anatomical information to the model, by learning affinities across point neighborhoods.
Our method is competitive on the \textsl{MSD} and \textsl{LiTS} public datasets in Dice coefficient and average surface distance scores compared to four pioneering point-based methods.
\camera{Our code is available at {\small \href{https://github.com/ZhangXiaotong015/GrPn}{https://github.com/ZhangXiaotong015/GrPn}}.}


\end{abstract}

\section{Introduction}
\label{sec:intro}
\begin{figure}[b]
  \centering
  \begin{tabular}{c}
    \includegraphics[width=.98\linewidth]{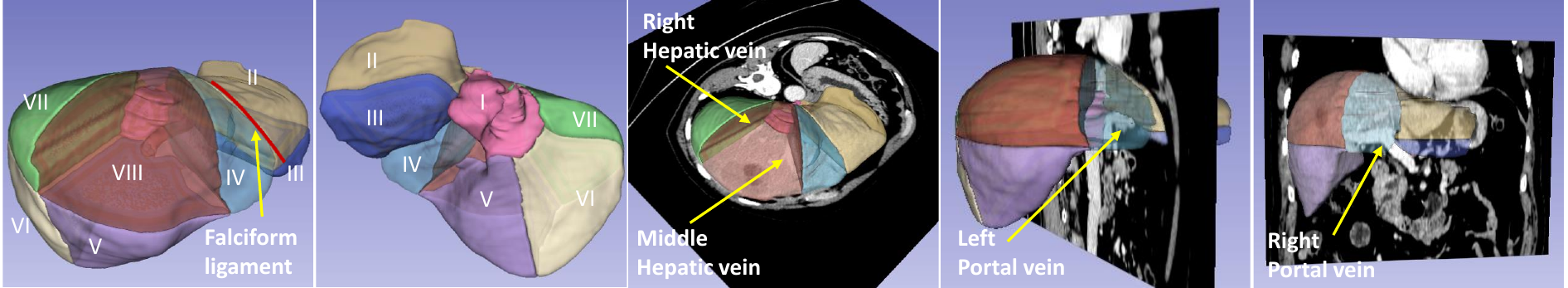} %
  \end{tabular}
  \caption{
    Couinaud segmentation is challenging because it requires prior knowledge of the liver vessels. 
    \camera{I--VIII indicate different Couinaud segments.}
  }
  \label{fig:summary}
\end{figure}
Effective treatment for primary liver cancer relies on two main procedures: liver resection and radiofrequency ablation \cite{b1}.
Both approaches depend on accurate Couinaud segmentation, to reduce the risk of main vessel puncture, and to guide the placement of ablation needles \cite{b2,b3}.
Couinaud segmentation divides the liver into eight functionally independent segments.
The right-, middle- and left-hepatic veins divide the liver into four sections. 
These sections are then further split by the horizontal plane defined by the portal vein, as shown in \fig{summary}.

\update{Prior work for automatically Couinaud segmentation, voxelizes the liver CT to be used in $3$D convolutional neural networks (CNNs) \cite{b34,b6,b7,b35}.
More recently \cite{b8}, computes point embeddings from sampled $3$D points within the liver area}.
Using sample $3$D points has the added value that they preserve the physical CT resolution, without the need to resize or crop along the axial direction. 
Therefore, here we restrict our focus to point-based methods.
\update{While relying on $3$D point-based representations, our proposed method does not need prior liver vessel information, unlike 
 Zhang et al. \cite{b8}}.
Yet, without prior liver vessel knowledge, we lose anatomical information. 
To incorporate this anatomical information, we add a graph reasoning module, learning affinities between the $3$D point embeddings.

To summarize:
(i) We propose a $3$D point-based method for Couinaud segmentation that \textsl{removes the need for prior knowledge} of the liver vessel structure;
(ii) We \textsl{extend $2$D graph reasoning to a $3$D version}, and use it to learn affinities between points along the liver, thus, adding implicitly anatomical liver structure; 
(iii) We evaluate on two public datasets: \textsl{MSD} and \textsl{LiTS} and \textsl{show competitive accuracy} when compared to four popular point-based segmentation methods.

\section{Related work}
\noindent\textbf{Couinaud segmentation.}
Prior work on automatic Couinaud segmentation creates liver atlases \cite{b5,b4}, divides the liver into voxels to be used in $3$D convolutional neural networks (CNNs) \cite{b34,b6,b7,b35}, or builds deep models on top of sampled $3$D points \cite{b8}.
Atlas-based and partial CNN-based methods \cite{b34,b5,b4,b35} require manual landmarks along the hepatic veins, whereas the other prior-free CNN-based methods \cite{b6,b7} on CT images need to resize the CT volume to a fixed grid size which changes the physical resolution of the CT images. 
Point-based models \cite{b8} address the limitation of voxelized methods, while still requiring prior liver vessel information.
Here, we build on $3$D point-based methods, while discarding the need for prior anatomical information, and learning this implicitly via dynamic graph reasoning.

\smallskip\noindent\textbf{Dynamic graph reasoning.}
Dynamic graph reasoning is widely used in both image-based \cite{b21,b23,b22,b12} as well as point-based semantic segmentation methods \cite{b24,b26,b25}, to capture long-range dependencies.
Most image-based methods \cite{b21,b23,b22} consider all position pairs when calculating affinities, resulting in high complexity.
Unlike these methods, DGMN \cite{b12} proposed an adaptive sampling method that considers only limited positions.
Similarly, point-based methods also suffer from high complexity of affinity calculation.
$K$-nearest neighbor (k-NN) is typically used for the complexity reduction, as in \cite{b26,b25}. 
Alternatively, Ma \etal \cite{b24} propose to learn channel dependencies instead of dependencies between nodes to capture global contextual information while reducing the computations.
Here, we take advantage of both k-NN in the point domain, and the adaptive sampling method in \cite{b12} to reduce computations.

\section{Couinaud liver segmentation}
\label{sec:method}
\begin{figure}[t]
  \centering
  \begin{tabular}{c}
    \includegraphics[width=1\linewidth]{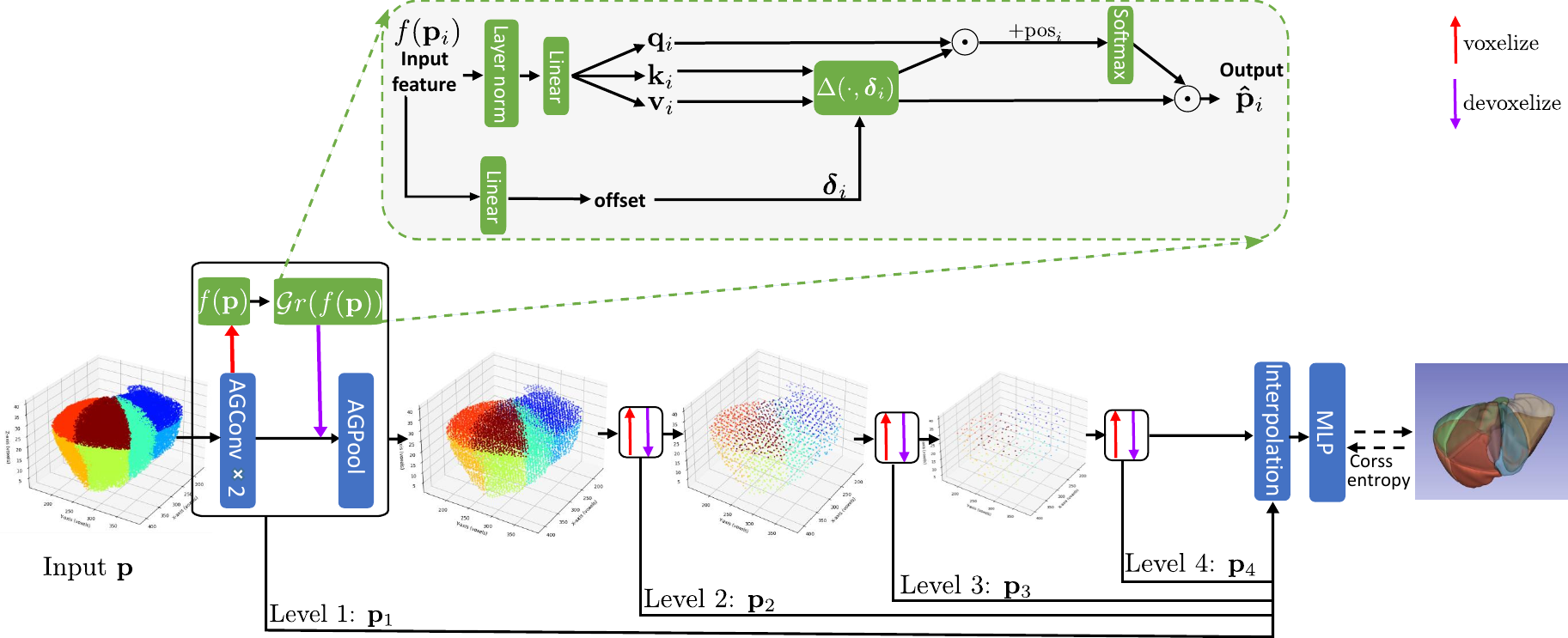} %
  \end{tabular}
  \caption{
    Network architecture. 
    We build on the design of \cite{b9}, and extend this with the green blocks: 
    grid feature embeddings, $f(\mathbf{p})$ adding anatomical information; 
    and graph reasoning model, $\mathcal{G}r(f(\mathbf{p}))$, learning dynamic affinities in neighborhood areas.
  }
  \label{fig:network}
\end{figure}
Our model starts from a set of $3$D points, $\mathbf{p}$, sampled from the liver region over the complete CT volume, and their associated intensities. 
We follow the design of Adaptive Graph CNN (AGCNN) \cite{b9}, processing the points at four levels, as in \fig{network}.
We extend AGCNN with the green blocks: the grid feature embeddings, $f(\mathbf{p})$, enhancing the anatomical information; and graph reasoning model, $\mathcal{G}r(f(\mathbf{p}))$, which dynamically learns affinities across points.  
At the last level, we interpolate the point embeddings, $\mathbf{p}_4$, and feed the result to an MLP with two layers ($\{64,8\}$). 
We predict the eight Couinaud segments, and use the cross-entropy loss to train the model.
\subsection{Relation to Adaptive Graph CNN}
 \begin{figure}[b]  
     \centering
     \includegraphics[width=0.75\textwidth]{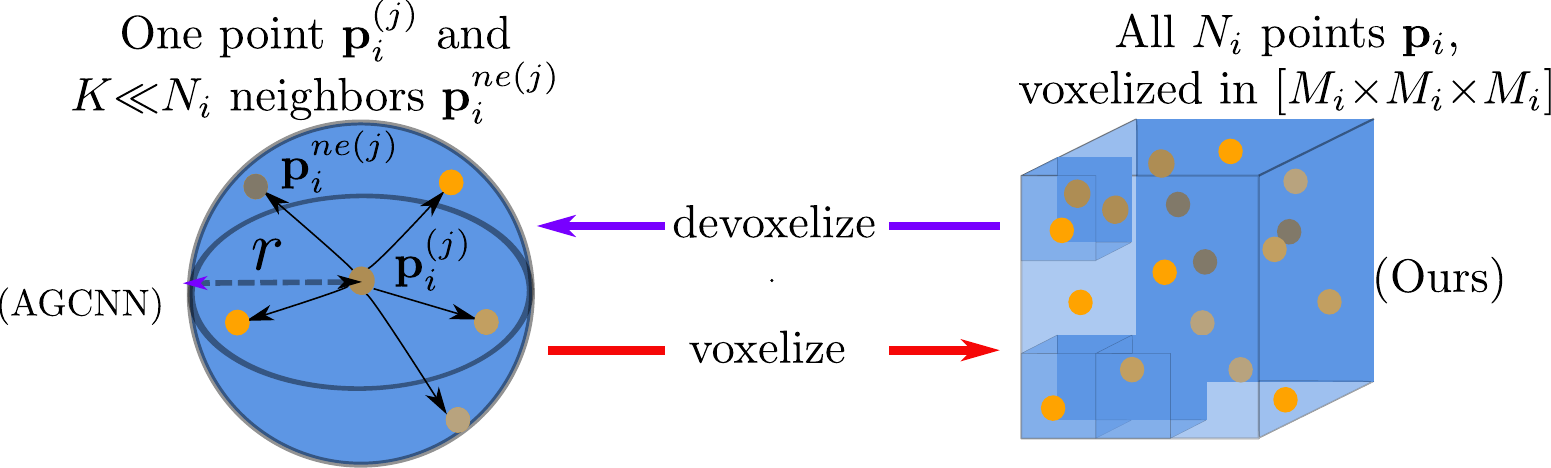}
     \caption{
     \textbf{Left:} AGCNN computes offline a set of $K{\ll}N_i$ neighbors $\mathbf{p}_i^{ne(j)}$ for each point indexed by $j$, $\mathbf{p}_i^{(j)}$ in level $i$.
     \textbf{Right:} Our approach voxelizes all $N_i$ points, $\mathbf{p}_i$ in level $i$, into a grid $[M_i{\times}M_i{\times}M_i]$, and dynamically focuses on $3^3$ voxels when learning affinities.    
     }
     \label{fig:vd}
\end{figure}

At the first level, AGCNN starts from a set of $3$D points, $\mathbf{p}_1$, and their associated CT intensities. 
Before training, AGCNN precomputes a set of $K{\ll}N_{i}$ neighbors for each point at different levels $i{\in}\{1,2,3,4\}$, where $N_i$ is the number of points at level $i$.
We compute the $K$ neighbors using a ball-query sampling scheme, with radius $r$, as in \fig{vd}. 
At the first level the neighbors of point $j$, $\mathbf{p}_1^{ne(j)}$, are taken from the initial set of points $\mathbf{p}_1$, while at lower resolution levels, the neighbors are taken from the previous level points, $\mathbf{p}_{i-1}, i{\in}\{2,3,4\}$.

To add implicit anatomical vessel structure to AGCNN, we use a graph reasoning module $\mathcal{G}$.
Specifically, at each level $i$ we voxelize all \update{$N_{i}$ points, \update{$\mathbf{p_{i}}$} into a grid of size $[M_{i}{\times}M_{i}{\times}M_{i}]$, using the method of Liu \etal \cite{b10}}.
In this voxelized space, the graph reasoning module $\mathcal{G}r(\cdot)$ dynamically learns the informative voxels.
And it selects $3^3$ voxels with which to compute affinities.
Intuitively, AGCNN computes affinities of points in the current level with the previous level, while our graph reasoning module computes affinities among the current level points.

\subsection{Graph-based implicit anatomical information}
\noindent\textbf{Grid feature embeddings.} 
\update{The inputs to the model are a set of $3$D points in $\mathbb{R}^3$, together with their corresponding CT intensity values.}
We revoxelize the point features and obtain the voxelized point embeddings, $f(\mathbf{p}_i)$ at each level, $i$, by performing two $3$D residual convolutions.

\smallskip\noindent\textbf{Dynamic graph reasoning.} 
\update{The grid-space $[M_{i}{\times}M_{i}{\times}M_{i}]$ at each level $i$ contains many voxels $\left({\geq}\left(\frac{32}{2^{i-1}}\right)^3\right)$ at different levels}.
This causes memory bottlenecks when calculating the affinity matrix.
To address this, we draw inspiration from DGMN \cite{b12} and extend their method from $2$D to $3$D.
Specifically, we consider only a subset of $3^3\camera{(=27)}$ voxels out of the $[M_{i}{\times}M_{i}{\times}M_{i}]$ grid, when computing affinities.
Following DGMN \cite{b12}, we use self-attention \cite{b13} to define affinities between voxelized points.
Given the voxelized point feature $f(\mathbf{p}_i)$ at level $i$, we first project this to \textsl{query} $\mathbf{q}_i$, \textsl{key} $\mathbf{k}_i$, and \textsl{value} $\mathbf{v}_i$, via a shared linear layer.
Additionally, we extend the offsets to learnable $3$D offsets $\boldsymbol{\delta}_i$ pointing to a set of $3^3$ voxels. 
These $3^3$ voxels, to which the offsets are pointing, should contain all the useful anatomical information encoded by neighboring voxels.
Similar to DGMN \cite{b12}, we use a deformable unfold layer \cite{b11}, $\Delta(\cdot, \boldsymbol{\delta}_i)$, to adapt the keys and values --- $\mathbf{k}_i$, $\mathbf{v}_i$.
The output of the graph reasoning is simply a $3$D self-attention block over deformed keys and values, as shown in \fig{network}:
\begin{align}
    \hat{\mathbf{p}}_i = \text{softmax}(\mathbf{q}_i \cdot \Delta(\mathbf{k}_i, \boldsymbol{\delta}_i) + \text{pos}_i) \cdot \Delta(\mathbf{v}_i, \boldsymbol{\delta}_i),
    \label{eq:atten}
\end{align}
\update{where $\boldsymbol{\delta}_i$=$\text{Linear}(f(\mathbf{p}_i))$, $\boldsymbol{\delta}_i{\in}\mathbb{R}^{3\times3^3}$, $\text{pos}_i{\in}\mathbb{R}^{3^3}$ are the positional embeddings for the query $\mathbf{q}_{i}$ of input features \cite{b14}}. 
Finally, we devoxelize $\hat{\mathbf{p}}_i$ back to point representations using the coordinate-based interpolation \cite{b10}.


\section{Experiments on Couinaud segmentation}
\label{sec:exp}

\noindent\textbf{Dataset description.}
We evaluate our method on two public datasets: \textsl{MSD} \cite{b15} and \textsl{LiTS} \cite{b16}.
Given that there are no Couinaud segment annotations in these two datasets, we use the annotations of Tian \etal \cite{b7} and Zhang \etal \cite{b8}.
\update{The \textsl{MSD} and \textsl{LiTS} datasets contain $192$ and $131$ annotated CT scans, respectively, with in-plane resolutions of $0.57$--$0.98$ mm and $0.56$--$1.00$ mm, and interplanar resolutions of $1.25$--$7.50$ mm and $0.70$--$5.00$ mm.}

\smallskip\noindent\textbf{Implementation details.}
We reoriented all CT scans to left-posterior-inferior (LPI) and maintain the original CT image origin and spacing for all experiments.
We divide both datasets into training\slash validation\slash test sets following the ratio $10/3/7$.
The CT values are truncated to the range of $[-100,300]$ Hounsfield units and then normalized to $[0,1]$.
We consider each voxel, $\mathbf{v}$, of a CT scan as a point, and compute the point coordinates as: $\mathbf{p}{=}sd \mathbf{v}{+}o$ where $s$, $d$ and $o$ are the physical spacing, direction and origin parameters recorded in the CT.
We also normalize the physical coordinates of the points to $[0,1]$. 
The $r$ in the ball-query sampling (\fig{vd}) is equal to $\frac{1}{2\cdot64}$ and $\frac{1}{2\cdot32}$, \update{and the first-scale grid size is $64^3$, and $32^3$} for the \textsl{MSD} and \textsl{LiTS} datasets, respectively.
\camera{The number of neighbors ($K{\le}100$) of a point depends on the $r$ used in ball-query sampling. 
We use the same point down-sampling ratio as AGCNN \cite{b9}.}
We randomly sample $10\%$ points (${\approx}$50K) for each training iteration.
We use $400$ epochs, the SGD optimizer with a momentum of $0.98$, and a learning rate of $0.01$.
\update{For all the experiments, we use an NVIDIA A100 (40GB) GPU.} 
We consider point-based baselines: PointNet \cite{b32}, PointNet${+}{+}$ \cite{b33}, AGCNN \cite{b9}, and Zhang \etal's \cite{b8} method, using their default settings and we follow their official implementation.

\smallskip\noindent\textbf{Evaluation metrics.}
We evaluate all methods only on the liver region, by masking out other areas.
\camera{The same liver masks were used for all experiments.} 
We report \textsl{Dice} coefficient and average surface distance (\textsl{ASD}) in our evaluation.
We use \textsl{Torchmetrics} \cite{b30} and \textsl{MONAI} \cite{b28} to calculate the metrics. \camera{These results are different from Zhang \etal \cite{b8}, because they use their own implementation for the metrics}. We also report inference times and GFLOPs \cite{calflops}.

\begin{table*}[t]
    \centering
    \caption{
        \textbf{Quantitative evaluation on \textsl{MSD}:} 
        We report results per segments (I -- VIII), as well as the average.
        Our method achieves the highest average in \textsl{Dice} coefficient, and comparable average in \textsl{ASD} with \textsl{PointNet}, demonstrating the effectiveness of the added implicit knowledge.
        \update{(We denote with $^*$ the use of extra vessel-priors.)}
    }
    \resizebox{1\linewidth}{!}{%
    \begin{tabular}{l ccccc ccccc}
    \toprule
            & \multicolumn{5}{c}{\textsl{MSD} (Dice \%) $\uparrow$} & \multicolumn{5}{c}{\textsl{MSD} (ASD mm) $\downarrow$}\\
    \cmidrule(lr){2-6}  
    \cmidrule(lr){7-11} 
            & PointNet & PointNet${+}{+}$ & AGCNN &  Zhang \etal & Ours &
            PointNet & PointNet${+}{+}$ & AGCNN &  Zhang \etal & Ours \\
            & \cite{b32} & \cite{b33} & \cite{b9} &  \cite{b8}* & &
            \cite{b32} & \cite{b33} & \cite{b9} &  \cite{b8}* &  \\
    \midrule
    (I) &62.26 &72.26 &63.96 &80.45 &\textbf{83.31} 
      &4.04 &5.01 &2.76 &2.96 &\textbf{1.97}\\
    (II) &81.53 &80.27 &77.79 &82.39 &\textbf{86.71} 
       &1.67 &\textbf{1.02} &6.19 &4.75 &1.89\\
    (III) &69.80 &72.75 &64.55 &75.06 &\textbf{79.35} 
        &3.31 &\textbf{1.96} &5.78 &3.30 &2.51\\
    (IV) &61.74 &68.04 &63.26 &69.83 &\textbf{73.26} 
       &5.76 &5.90 &6.52 &5.46 &\textbf{4.12}\\
    (V) &70.55 &71.78 &68.78 &71.49 &\textbf{75.89} 
      &\textbf{3.56} &5.61 &11.75 &5.91 &6.36\\
    (VI) &75.88 &75.29 &69.91 &72.51 &\textbf{79.51} 
       &\textbf{2.68} &4.57 &8.20 &5.23 &5.15\\
    (VII) & 82.38 &82.24 &80.35 &80.56 &\textbf{85.11} 
        &\textbf{2.47} &3.02 &5.62 &5.57 &3.27\\
    (VIII) &75.86 &76.03 &73.72 &75.74 &\textbf{{80.16}} 
         &\textbf{4.10} &4.19 &6.79 &4.82 &4.84\\ \midrule
    Avg &72.50 &74.83 &70.29 &76.00 &\textbf{80.41} 
        &\textbf{3.45} &3.91 &6.70 &4.75 &3.76\\   
    \midrule
    & \multicolumn{5}{c}{\update{Time (s) per case}} & \multicolumn{5}{c}{\update{\textsl{GFLOPs}}}\\
    \cmidrule(lr){2-6}  
    \cmidrule(lr){7-11} 
    & 2.81 & 21.55 & 6.13 & 1.73 & 10.03 
    & 302.38 & 64.65 & 638.54 & 608.85 & 771.15 \\    
    \bottomrule
    \end{tabular}}
    \label{tab:results_MSD}
\end{table*}

\subsection{Quantitative evaluation}
\tab{results_MSD} and \tab{results_LiTS} show the quantitative comparison in \textsl{Dice} coefficient and \textsl{ASD} scores.
In \tab{results_MSD}, our method achieves the highest \textsl{Dice} score for each Couinaud segment.
Both our method and \textsl{PointNet} have lower \textsl{ASD} scores compared to the other methods.
\textsl{PointNet} has a low \textsl{ASD} score because it uses a voting scheme at inference to reduce false positives.
However, both \textsl{PointNet}${++}$ and \textsl{PointNet} have large segmentation errors for segment (I).
This may be due to the segment having the lowest volume fraction in the liver.
In \tab{results_LiTS}, our method has the highest \textsl{Dice} coefficient and the lowest \textsl{ASD} averaged over all segments.
This is especially positive, given the high axial resolution (\ie ${\le}1.0$ mm) of the \textsl{LiTS} dataset, which increases the voxel-wise class imbalance.

\update{PointNet++, AGCNN, and our method use multi-scale point sampling.
This results in longer inference times compared to PointNet and Zhang \etal, which do not employ multi-scale point sampling.}
\update{The GFLOPs of AGCNN and our method depend on the radius used in the ball-query sampling.}

\begin{table*}[t]
    \centering
    \caption{
        \textbf{Quantitative evaluation on \textsl{LiTS}:} 
        Results per Couinaud segment (I -- VIII) and the overall average.
        Our method, without prior vessel knowledge, achieves the highest average in \textsl{Dice} coefficient and the lowest average in \textsl{ASD}. 
    }
    \resizebox{1\linewidth}{!}{%
    \begin{tabular}{l ccccc ccccc}
    \toprule
            & \multicolumn{5}{c}{\textsl{LiTS} (Dice \%) $\uparrow$} & \multicolumn{5}{c}{\textsl{LiTS} (ASD mm) $\downarrow$}\\
    \cmidrule(lr){2-6}  
    \cmidrule(lr){7-11} 
            & PointNet & PointNet++ & AGCNN &  Zhang \etal & Ours &
            PointNet & PointNet++ & AGCNN &  Zhang \etal & Ours \\ 
            & \cite{b32} & \cite{b33} & \cite{b9} & \cite{b8}* & &
            \cite{b32} & \cite{b33} & \cite{b9} &  \cite{b8}* & \\
    \midrule
    (I) &49.80 &37.85 &69.20 &\textbf{73.64} &68.84 
      &8.39 &8.58 &\textbf{4.36} &6.54 &6.97\\
    (II) &70.69 &72.78 &82.62 &82.82 &\textbf{86.17}
       &9.39 &5.31 &4.58 &4.81 &\textbf{3.18}\\
    (III) &58.26 &65.68 &76.09 &72.46 &\textbf{80.82}
        &15.18 &7.42 &6.04 &5.09 &\textbf{3.18}\\
    (IV) &53.87 &70.19 &73.88 &\textbf{75.40} &75.24
       &8.46 &9.69 &10.00 &\textbf{7.88} &8.38\\
    (V) &80.46 &80.51 &78.97 &81.92 &\textbf{83.03}
      &\textbf{5.04} &5.85 &7.25 &6.49 &6.20\\
    (VI) &77.69 &79.23 &73.45 &79.28 &\textbf{79.29}
       &4.55 &6.02 &5.76 &6.67 &\textbf{3.78}\\
    (VII) &79.92 &81.71 &80.42 &\textbf{83.40} &82.79
        &\textbf{4.38} &4.93 &6.75 &4.63 &4.60\\
    (VIII) &77.20 &79.29 &77.53 &80.02 &\textbf{80.26}
         &6.45 &6.67 &7.71 &\textbf{5.84} &7.71\\ \midrule
    Avg &68.49 &70.90 &76.52 &78.62 &\textbf{79.56}
        &7.73 &6.81 &6.56 &6.00 &\textbf{5.50}\\    
    \midrule
    & \multicolumn{5}{c}{\update{Time (s) per case}} & \multicolumn{5}{c}{\textsl{\update{GFLOPs}}}\\
    \cmidrule(lr){2-6}  
    \cmidrule(lr){7-11} 
    & 10.62 & 106.97 & 19.81 & 6.52 & 13.83 
    & 302.38 & 64.65 & 256.10 & 608.85 & 141.56 \\    
    \bottomrule
    \end{tabular}}
    \label{tab:results_LiTS}
\end{table*}

\begin{figure}[t]
  \centering
  \begin{tabular}{c}
    \includegraphics[width=1\linewidth]{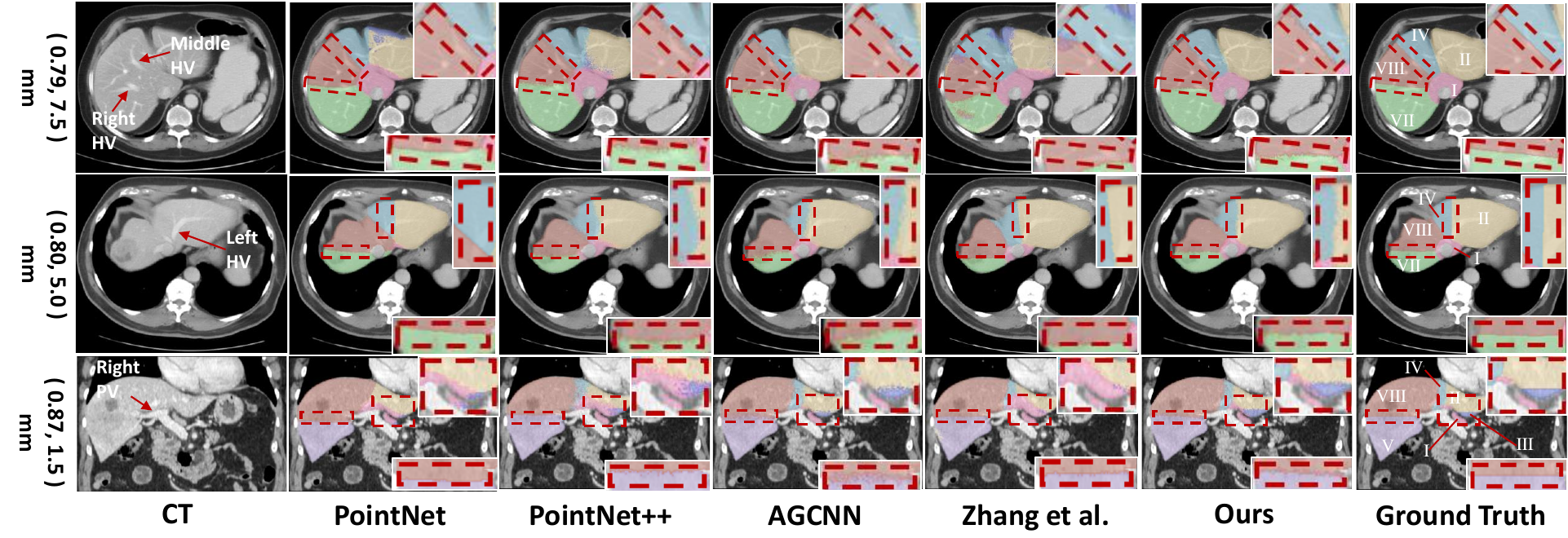} %
  \end{tabular}
  \caption{
    \textbf{Qualitative evaluation on \textsl{MSD}.} 
    The anatomical landmarks \cite{b29} are marked with red arrows in the CT image, as a reference.
    The red dotted bounding-box highlights the boundary between segments. 
    Our method predicts segment boundaries that are closest to the ground truth.
    We show to the left the in-plane and interplane resolutions. (HV: heaptic vein, PV: portal vein)
  }
  \label{fig:viz_MSD}
\end{figure}
\begin{figure}[t]
  \centering
  \begin{tabular}{c}
    \includegraphics[width=1\linewidth]{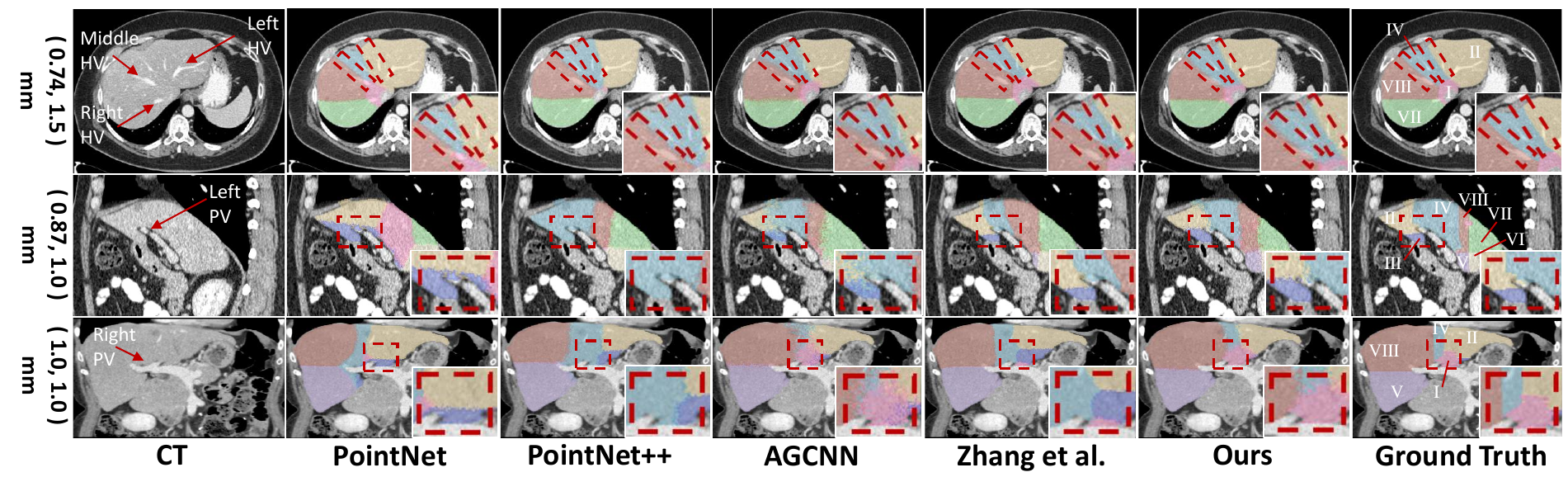} %
  \end{tabular}
  \caption{
    \textbf{Qualitative evaluation on \textsl{LiTS}.} 
    Three cases in axial, sagittal, and coronal views.
    The red arrows mark the anatomical landmarks \cite{b29}, and the bounding-boxes highlight segment boundaries.
    Our method displays the most accurate boundaries.
  }
  \label{fig:viz_LiTS}
\end{figure}

\subsection{Qualitative evaluation}
We visualize the Couinaud segmentations on the \textsl{MSD} and \textsl{LiTS} datasets, as shown in \figs{viz_MSD}{viz_LiTS}.
In \fig{viz_MSD}, we show three different cases with varying axial spacing from $1.5$ mm to $7.5$ mm, in the \textsl{MSD} dataset.
For the cases with lower axial resolution ($5$ mm and $7.5$ mm), we show the plane in axial view, and use the red dotted bounding-box to highlight the boundaries between segments. 
We also show one case with relatively high axial resolution ($1.$5 mm) in a coronal view, in the last row of \fig{viz_MSD}.
All boxes are on the same location, in the corresponding CT image.
On the last row, both our method and Zhang \etal's \cite{b8} boundaries follow the box centerline. 
However, unlike Zhang \etal's \cite{b8}, our method correctly recognizes segment (III) in this case, as seen in the right-upper corner.

In \fig{viz_LiTS}, we show the boundary comparison for three cases in three views (axial, coronal and sagittal).
Similar to the results in \fig{viz_MSD}, the boundaries of our segments are located on the box centerline. 
In the second and third rows, ours method show the most precise precictions.

\begin{table*}[t]
    \centering
    \caption{
    \textbf{Model ablation study in \textsl{Dice} coefficient and average surface distance (\textsl{ASD}) on the \textsl{MSD} \cite{b15} and \textsl{LiTS} \cite{b16} datasets}.
    (a) AGCNN \cite{b9} baseline; (b) without graph reasoning, $\mathcal{G}r(f(\mathbf{p}))$, in our model; (c) without grid feature embeddings, $f(\mathbf{p})$, in our proposed model; (d) our proposed model.
    All our model components contribute to the final model’s scores.
    }
    \label{tab:ablation_metrics}
    \resizebox{0.65\linewidth}{!}{ 
        \begin{tabular}{lcccc}
        \toprule
         & \multicolumn{2}{c}{\textsl{MSD}} & \multicolumn{2}{c}{\textsl{LiTS}} \\
        \cmidrule(lr){2-3} 
        \cmidrule(lr){4-5} 
        & Dice $\uparrow$ (\%) & ASD $\downarrow$ (mm) & Dice $\uparrow$ (\%) & ASD $\downarrow$ (mm) \\
        \midrule
        (a) & 70.29 ($\pm$ 16.69) & 6.70 ($\pm$ 5.80) & 76.52 ($\pm$ 8.90) & 6.56 ($\pm$ 2.65) \\ 
        (b) & 78.02 ($\pm$ 11.06) & 4.00 ($\pm$ 3.39) & 77.46 ($\pm$ 10.87) & 6.58 ($\pm$ 4.03) \\ 
        (c) & 68.98 ($\pm$ 14.58) & 7.73 ($\pm$ 5.19) & 72.67 ($\pm$ 10.30) & 9.81 ($\pm$ 5.44) \\ 
        (d) & \textbf{80.41} ($\pm$ 10.74) & \textbf{3.76} ($\pm$ 4.04) & \textbf{79.56} ($\pm$ 6.95) & \textbf{5.50} ($\pm$ 2.42) \\ 
        \bottomrule
        \end{tabular}
    }
\end{table*}

\subsection{Model ablation experiments}
To verify that all the components of our model contribute to the segmentation, we perform model ablations in \tab{ablation_metrics}.
We consider four cases: (a) the baseline AGCNN \cite{b9}; (b) our method with grid feature embeddings, $f(\mathbf{p})$, but without graph reasoning, $\mathcal{G}r(f(\mathbf{p}))$; (c) our model with graph reasoning, $\mathcal{G}r(f(\mathbf{p}))$, but without grid feature embeddings, $f(\mathbf{p})$; (d) our complete model.   
Grid feature embeddings $f(\mathbf{p})$ are an important bridge between features in $3$D point-space, $\mathbf{p}$, and the graph reasoning module $\mathcal{G}r(f(\mathbf{p}))$.
As seen for the setting (c) in \tab{ablation_metrics}, the graph reasoning module fails to work directly, on the voxelized point-features.
In addition, the graph reasoning module in itself contributes to the model predictions, as seen in the setting (b) of \tab{ablation_metrics}. 
Removing $\mathcal{G}r(f(\mathbf{p}))$ decreases the dice scores by $\approx 2 \%$ on both the \textsl{MSD} and \textsl{LiTS} datasets.
On both the \textsl{LiTS} data and the \textsl{MSD} dataset, all model components prove useful.

\begin{figure}[t!]
    \centering
    \includegraphics[width=.8\textwidth]{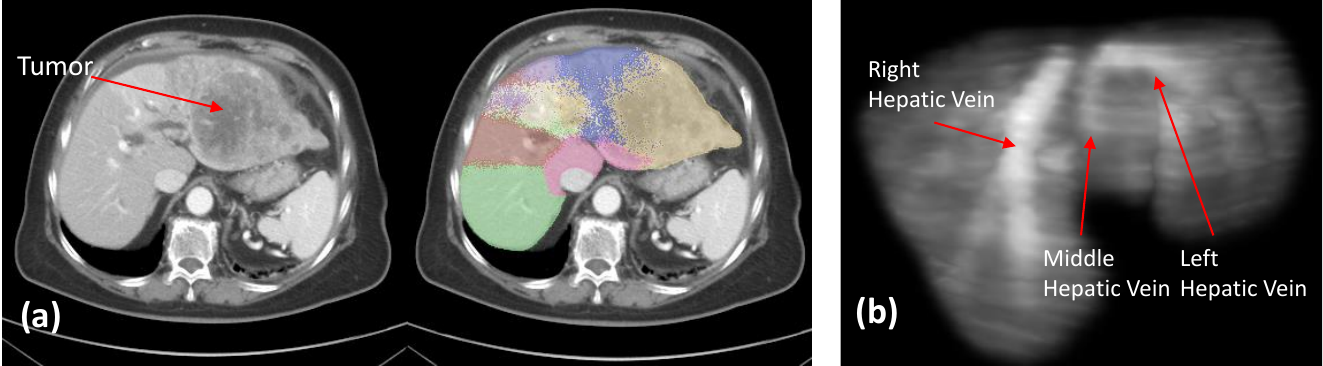}
    \caption{\small
    \textbf{Failure case analysis.} (a) The anatomical landmarks disappear in the tumor region, marked by the red arrow. (b) Output of $\mathcal{G}r(f(\mathbf{p}))$ of a successful case. The model learns the graph affinities based on the anatomical landmarks in the CT images.
    }
    \label{fig:failure}
\end{figure}

\section{Discussion and conclusion}
\label{sec:disc_con}
The proposed method does not need explicit vessel priors, unlike \cite{b8}. 
Moreover, as shown in \fig{failure} (b), it can learn anatomical liver landmarks in the graph $\mathcal{G}r(f(\mathbf{p}))$.
However, the proposed method cannot accurately discriminate segments in the liver when a large tumor appears, as shown in \fig{failure} (a).
Here, the partial anatomical landmarks in the CT image are absent, such as the left hepatic vein and the left portal vein.
To conclude, we propose a $3$D point-based method for Couinaud segmentation for CT images.
Our model incorporates implicitly anatomical information, by learning affinities between voxels in the CT volume in a dynamic graph reasoning module.
This implicit anatomical information makes our model competitive compared to prior point-based methods, where we exceed \textsl{PointNet}, \textsl{PointNet}${++}$, \textsl{AGCNN} and Zhang \etal \rq s \cite{b8} method on the \textsl{MSD} and \textsl{LiTS} public benchmarks. 
This approach avoids the time-consuming definition of the prior vessel structure in the liver, while still showing competitive performance.

\smallskip\noindent\textbf{Acknowledgements.} This work was supported by China Scholarship Council under Grant 202108310010.

\bibliographystyle{splncs04}
\bibliography{bibliography}
\end{document}